\documentclass[sigconf]{acmart}

\AtBeginDocument{%
  \providecommand\BibTeX{{%
    \normalfont B\kern-0.5em{\scshape i\kern-0.25em b}\kern-0.8em\TeX}}}

\setcopyright{acmcopyright}
\copyrightyear{2018}
\acmYear{2018}
\acmDOI{XXXXXXX.XXXXXXX}
\usepackage{enumitem}
\usepackage{multirow}
\usepackage{array}
\usepackage{booktabs} 
\usepackage{microtype}
\usepackage{times}
\usepackage{latexsym}
\usepackage{multirow}
\usepackage{graphics}
\usepackage{booktabs} 
\usepackage{bm}
\usepackage{url}
\usepackage{mathrsfs}
\usepackage{multirow}
\usepackage{balance}
\usepackage{amsthm}
\usepackage{amsmath}
\usepackage{amstext}
\usepackage[ruled,vlined,linesnumbered]{algorithm2e}
\usepackage{subcaption}
\usepackage{overpic}
\usepackage{color}
\usepackage{enumitem}
\usepackage{array}
\usepackage{booktabs} 
\usepackage{enumitem}
\usepackage{xcolor}

\begin{document}

\title{Shortcut Learning of Large Language Models in Natural Language Understanding}

\author{Mengnan Du}
\affiliation{%
  \institution{New Jersey Institute of Technology}
  \city{Newark}
  \state{NJ}
  \country{USA}
}
\email{mengnan.du@njit.edu}

\author{Fengxiang He}
\affiliation{%
  \institution{JD Explore Academy}
  \city{Beijing}
  \state{Beijing}
  \country{China}
}
\email{fengxiang.f.he@gmail.com}

\author{Na Zou}
\affiliation{%
  \institution{Texas A\&M University}
  \city{College Station}
  \state{TX}
  \country{USA}
}
 \email{nzou1@tamu.edu}

\author{Dacheng Tao}
\affiliation{%
 \institution{The University of Sydney}
 \city{Sydney}
  \country{Australia}
 }
 \email{dacheng.tao@gmail.com}

\author{Xia Hu}
\affiliation{%
  \institution{Rice University}
  \city{Houston}
  \state{TX}
  \country{USA}
  }
  \email{xia.hu@rice.edu}


\begin{abstract}
 Large language models (LLMs) have achieved state-of-the-art performance on a series of natural language understanding tasks. However, these LLMs might rely on dataset bias and artifacts as shortcuts for prediction. This has significantly affected their generalizability and adversarial robustness. In this paper, we provide a review of recent developments that address the shortcut learning and robustness challenge of LLMs. We first introduce the concepts of shortcut learning of language models. We then introduce methods to identify shortcut learning behavior in language models, characterize the reasons for shortcut learning, as well as introduce mitigation solutions. Finally, we discuss key research challenges and potential research directions in order to advance the field of LLMs.
\end{abstract}

\maketitle

\section{Introduction}

Natural language understanding (NLU) is a subfield of artificial intelligence that requires computer software to comprehend input in the form of sentences. Representative NLU tasks include natural language inference (NLI), question answering (QA), reading comprehension, etc. Furthermore, NLU has a number of real-world applications, including Amazon Alexa, Siri, and Google Assistant. The major characteristic of NLU tasks is that they are difficult and typically require world knowledge and commonsense reasoning. Recently, large language models (LLMs), such as BERT~\cite{devlin2018bert}, RoBERTa~\cite{liu2019roberta}, T5~\cite{raffel2020exploring}, GPT-3~\cite{brown2020language}, have been reported to achieve state-of-the-art performance in a series of high-level NLU tasks. The LLM performance has even been reported to be significantly higher than human performance. However, the superior performance has only been observed in the benchmark test data that have the same distribution as the training set. Recent studies indicate that these LLMs are not robust and that the models do not remain predictive when the distribution of inputs changes~\cite{niven2019probing,utama2020towards,du2021towards}. Specifically, these LLMs have low generalization performance when applied to out-of-distribution (OOD) test data and are also vulnerable to various types of adversarial attack.  This leaves us wondering: \emph{Why are these LLMs not robust? Have these LLMs mastered the high-level semantic understanding and reasoning that we expect of them?  }

\begin{figure*}
  \centering
  \includegraphics[width=0.99\linewidth]{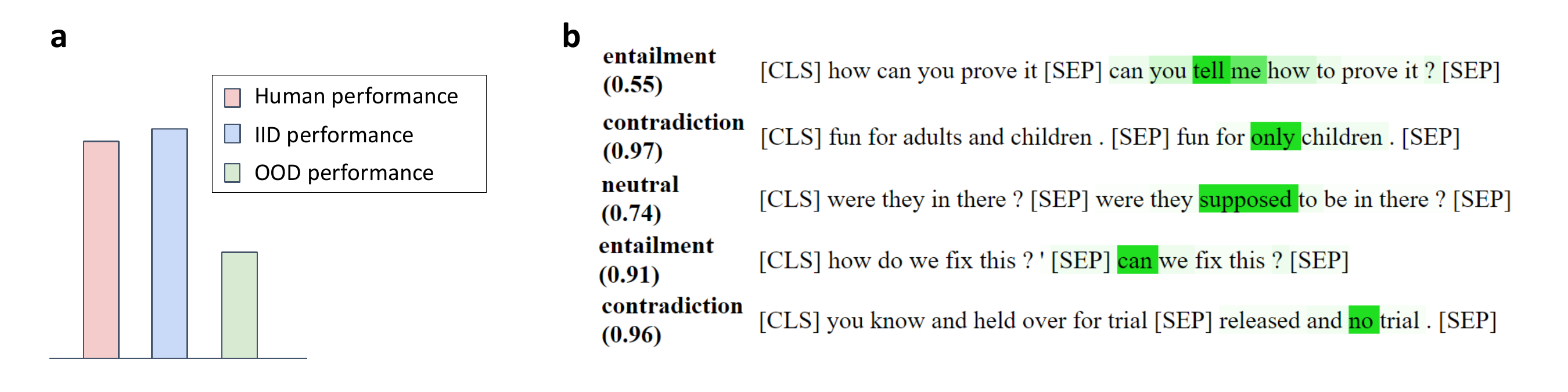}
  \caption{Shortcut learning behavior and its negative impact, taking natural language inference (NLI) task for example. The goal of NLI is to infer whether the relationship between two branches of input, i.e., premise and hypothesis, is entailment, contradiction, or neutral. (a) LLMs outperforms human performance for IID benchmark test set, while achieve much lower generalization performance on OOD test set. (b) A key reason is that LLMs primarily rely on the lexical bias and other kinds of shortcuts for prediction.}
  \label{fig:examples}
\end{figure*}

A major reason for the low robustness of LLMs is \textbf{shortcut learning}. The shortcut learning behavior has also been called other names in the literature, such as learning bias, superficial correlations, right for wrong reasons, Clever Hans effect\footnote{The eponymous horse appeared to be capable of performing simple intellectual tasks, but actually relied on involuntary cues given by its handler.}, etc. The shortcut learning behavior has been observed for a series of NLU tasks. For example, recent empirical analysis indicates that the performance of BERT-like models for the NLI task could be mainly explained by relying on spurious statistical cues such as unigrams `not', `do', `is' and bigrams `will not' (see Figure~\ref{fig:examples} (b))~\cite{niven2019probing,gururangan2018annotation}. Similarly, for the reading comprehension task, the models rely on the lexical matching of words between the question and the original passage, while ignoring the designed reading comprehension task~\cite{lai2021machine}. The current standard approach to training LLM is to use empirical risk minimization (ERM) on NLU datasets that typically contain various types of artifacts and biases. As such, LLMs have learned to rely on dataset artifacts and biases and capture their spurious correlations with certain class labels as shortcuts for prediction. The shortcut learning behavior has significantly affected the robustness of LLMs (see Figure~\ref{fig:examples} (a)), thus attracting increasing attention from the NLP community to address this problem.

In this work, we offer a comprehensive review of the shortcut learning problem in language models with a focus on medium-sized LLMs those typically having less than a billion parameters. The main emphasis is on the prevalent pre-training and fine-tuning paradigm utilized in NLU tasks (see Figure~\ref{fig:pre-training-fine-tuning}). We cover the concept of shortcut learning and robustness challenges in Section 2, detection approaches in Section 3, characterization of the corresponding reasons in Section 4, and mitigation approaches in Section 5. We also provide a further discussion of future research directions and connections with other directions in Section 6. Beyond the standard fine-tuning paradigm, in Section 7, we briefly discuss the challenges of shortcut learning posed by the prompt-based paradigm, especially regarding the massive language models which possess over a billion parameters, such as GPT-3 and T5.

\section{Shortcut Learning Phenomena}

\subsection{What is Shortcut Learning?}\label{sec:bias-types}
\begin{figure*}
  \centering
  \includegraphics[width=0.9\linewidth]{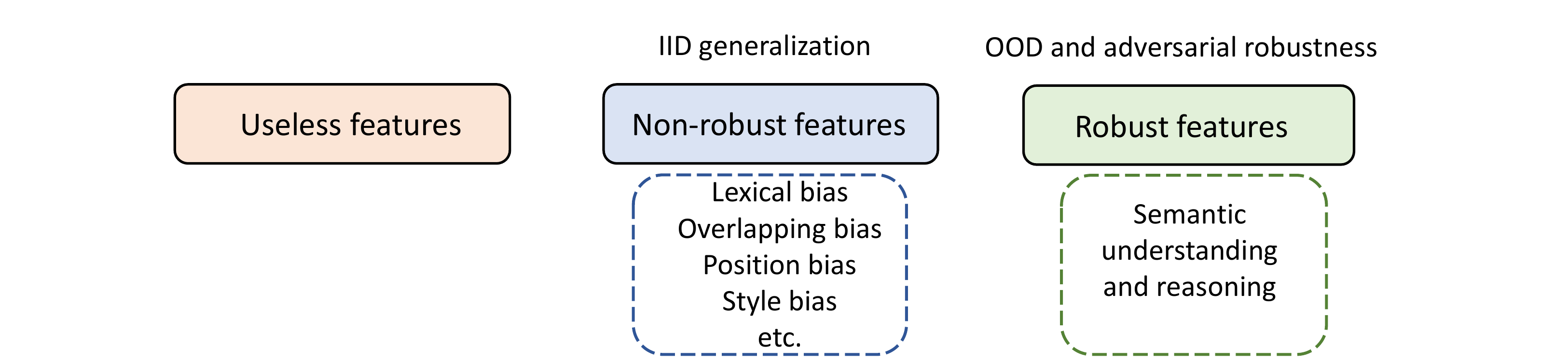}
  \caption{Features can be generally grouped into useless features, robust features, and non-robust features. Non-robust features indicate various kinds of biases captured by the model, which are not robust in the OOD setting. In contrast, robust features denote features of high-level semantic understanding that are robust to changes in the input.}
  \label{fig:robust-features}
\end{figure*}

Features captured by the model can be broadly categorized as useless features, robust features, and non-robust features
(see Figure~\ref{fig:robust-features}). Shortcut learning refers to the phenomenon that LLMs (especially those trained with standard ERM-based method) highly rely on non-robust features as shortcuts, failing to learn robust features and capture high-level semantic understanding and reasoning. Non-robust features do help generalization for development and test sets that share the same distribution with training data. However, they cannot generalize to OOD test sets and are vulnerable to adversarial attacks. Non-robust features are oriented from biases in the training data and come in different formats. In the following, we introduce several representative ones.
\begin{itemize}[leftmargin=*]
\item \emph{Lexical Bias}: Some lexical features have a high correlation of co-occurrence with certain class labels. These lexical features mainly consist of low-level functional words such as stop words, numbers, negation words, etc. A typical example is the NLI task, where LLMs are highly dependent on unintended lexical features to make predictions~\cite{niven2019probing,du2021towards}. For example, these models tend to give contradiction predictions whenever there exist negation words in the input samples, e.g., `never', `no'.

\item \emph{Overlap Bias}: It occurs in NLU applications with two branches of text, e.g., natural language inference, question answering, and reading comprehension. LLMs use the overlap of features between the two branches of inputs as spurious correlations as shortcuts. For example, reading comprehension models use the overlap between the passage and the question pair for prediction rather than solving the underlying task~\cite{lai2021machine}. Similarly, question answering models excel at test sets by relying on the heuristics of question and context overlap~\cite{sen2020models}.

\item \emph{Position Bias}: 
The distribution of the answer positions may be highly skewed in the training set for some applications. The LLMs would predict answers based on spurious positional cues. Take the question answering task for example,
the answers lie only in the k-th sentence of each passage~\cite{ko2020look}. As a result, question answering models rely on this spurious cue when predicting answers.

\item \emph{Style Bias}: 
The text style is a kind of pattern that is independent of semantics. Models have learned to rely on the erroneous text style as a shortcut rather than capturing the underlying semantics. Adversaries can use this style bias to launch adversarial attacks~\cite{qi-etal-2021-mind}.

\end{itemize}

\subsection{Generalization and Robustness Challenge }
The shortcut learning behavior could significantly hurt LLMs' \textbf{OOD generalization} as well as \textbf{adversarial robustness}. First, shortcut learning may result in significant performance degradation for OOD data.  
A common assumption is that training and test data are independently and identically distributed (IID). When LLMs are deployed in real-world applications with distribution shifts, this IID assumption will not hold any longer. 
These data typically do not contain the same types of bias and artifacts as the training data.
\begin{equation}
\small
\begin{aligned}
&\text {IID: } \boldsymbol{P}_{train}(\boldsymbol{X}, \boldsymbol{Y})=\boldsymbol{P}_{test}(\boldsymbol{X}, \boldsymbol{Y}) \\
&\text {OOD: } \boldsymbol{P}_{train}(\boldsymbol{X}, \boldsymbol{Y}) \neq \boldsymbol{P}_{test}(\boldsymbol{X}, \boldsymbol{Y})
\end{aligned}
\end{equation}
Using BERT-base as an example, there is a more than 20\% reduction in accuracy on the OOD test set compared to the accuracy on the in-distribution test sets for NLU tasks~\cite{du2021compressed}. To some extent, these models have solved the dataset rather than the underlying task. Second, shortcut learning produces models that are easily fooled by adversarial samples, which are generated when small and often imperceptible human-crafted perturbations are added to the normal input. One typical example is for the multiple choice reading comprehension task~\cite{si2019does}. BERT models are attacked by adding distracting information, resulting in a significant performance drop. Further analysis indicates that these models are highly driven by superficial patterns, which inevitably leads to their adversarial vulnerability.

\section{Shortcut Learning Detection}
In this section, we discuss methods to identify shortcut learning problems in NLU models.

\subsection{Comprehensive Performance Testing}
Traditional evaluations employ IID training-test split of data. The test sets are drawn from the same distribution as the training sets and thus share the same kind of biases as the training data. Models that simply rely on memorizing superficial patterns could perform acceptablely on the IID test set. This type of evaluation has failed to identify the shortcut learning problem. Therefore, it is desirable to perform more comprehensive tests beyond the traditional IID testing.

First, the OOD generalization test has been proposed as an alternative to the IID test.
Take the multi-genre natural language inference (MNLI) task for example, the HANS evaluation set is proposed to evaluate whether NLI models have syntactic heuristics: the lexical overlap heuristic, the subsequence heuristic, and the constituent heuristic~\cite{mccoy2019right}. Similarly, for the fact verification task, a symmetric test set is constructed that shares a philosophy similar to HANS~\cite{schuster2019towards}. These OOD tests have revealed dramatic performance degradation and exposed the shortcut learning problem of state-of-the-art LLMs.

Second, adversarial attacks could also be implemented to test the robustness of LLMs. For example, adversarial attacks have been used to reveal statistical bias in machine reading comprehension models~\cite{lai2021machine}.
Besides, the adversarial examples created through TextFooler~\cite{jin2020bert} are used to test the generalization of common sense reasoning models~\cite{branco-etal-2021-shortcutted-commonsense}. The results indicate that the models have learned non-robust features and fail to generalize towards the main tasks associated with the datasets.

Third, randomization ablation methods are proposed to analyze whether LLMs have used these essential factors to achieve effective language understanding. For example, word order is a representative one among these significant factors. Recent ablation results indicate that word order does not matter for pre-trained language models~\cite{sinha2021masked}. In particular, LLMs are pre-trained first on sentences with randomly shuffled word order and then fine-tuned on various downstream tasks. The results show that these models still achieve high accuracy. Similarly, another study~\cite{pham2020out} has observed that LLMs are insensitive to word order in a wide set of tasks, including the entire GLUE benchmark. These experiments indicate that LLMs have ignored the syntax when performing downstream tasks, and their success can almost be explained by their ability to model higher-order word co-occurrence statistics. 

\begin{figure*}
  \centering
  \includegraphics[width=0.9\linewidth]{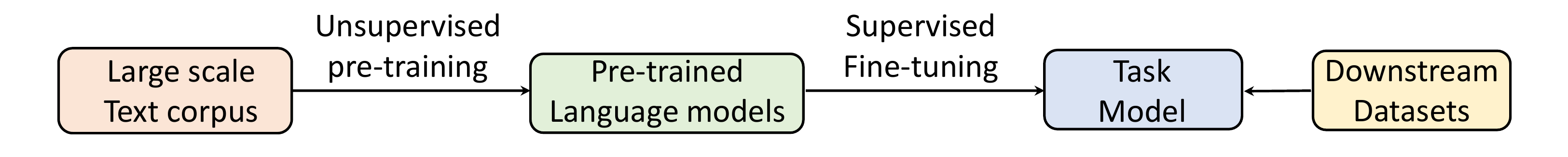}
  \caption{The pre-training then fine-tuning training paradigm. Shortcut learning can be attributed to different factors in this pipeline, including pre-trained language models, fine-tuning process, and downstream tasks.}
  \label{fig:pre-training-fine-tuning}
\end{figure*}

\subsection{Explainability Analysis}
Model explainability is another effective tool that the community has used to identify the shortcut learning problem. LLMs are usually considered black boxes, as their decision-making process is opaque and difficult for humans to understand. This presents challenges in identifying whether these models make decisions based on justified reasons or on superficial patterns. Explainability enables us to diagnose spurious patterns captured by LLMs.

The existing literature mainly employs the explanation in the format of feature attribution to analyze shortcut learning behavior in NLU models~\cite{du2021towards}. Feature attribution is the most representative paradigm among all explainability-based methods. In particular, for each token $x_i$ within a specific input $x$, the feature attribution algorithm $\psi$ will calculate the contribution score $\psi_i$, which denotes the contribution score of that token for model prediction.
For example, the Integrated Gradient~\cite{sundararajan2017axiomatic} interpretation method is used to analyze the model behavior of BERT-based models~\cite{du2021towards}. It is observed that LLMs rely on dataset artifacts and biases within the hypothesis sentence for prediction, including functional words, negation words, etc~\cite{du2021towards}. This shortcut learning behavior is summarized further using the long-tailed phenomenon. Specifically, the tokens in the training set could be modeled using a long-tailed distribution. The LLM models concentrate mainly on information on the head of the distribution, which typically corresponds to non-generalizable shortcut tokens. In contrast, the tail of the distribution is poorly learned, although it contains abundant information for an NLU task.

Beyond feature attribution, other types of explainability methods have also been used to analyze shortcut learning behaviors.
For example, instance attribution methods have been used to explain model prediction by identifying influential training data, which can be used to explain decision making logic for the current sample of interest~\cite{han2020explaining}. Empirical analysis indicates that the most influential training data share similar artifacts, e.g., high overlap between the premise and hypothesis for the NLI task. Furthermore, hybrid methods that combine feature attribution and instance attribution have also been used to identify artifacts in the data~\cite{pezeshkpour2021combining}. The resulting explanations have provided a more comprehensive perspective on the shortcut learning behavior of LLMs.

\section{Origins of Shortcut Learning}\label{sec:Characterization-for-Low-Robustness}
The problem of learning shortcuts in LLM models for NLU tasks is a result of multiple factors present in the training pipeline (see Figure~\ref{fig:pre-training-fine-tuning}). In this section, we will delve into these reasons and give particular emphasis to three key elements: the training datasets, the LLM model, and the fine-tuning training procedure.

\subsection{Skewed Training Dataset}
From a data standpoint, the NLU models' shortcut learning can be traced back to the annotation and collection artifacts of the training data to a large extent. Here, the training data include both the pretraining datasets as well as the downstream datasets (see  Figure~\ref{fig:pre-training-fine-tuning}). Training sets are typically built through the crowd-sourcing process, which has the advantage of being low-cost and scalable. However, the crowd-sourcing process results in collection artifacts, where the training data are imbalanced with respect to features and class labels. Furthermore, when crowd workers author parts of the samples, they produce certain patterns of artifacts, i.e., annotation artifacts~\cite{gururangan2018annotation,saxon2023peco}. Taking NLI as an example, the average sentence length of the hypothesis branch is shorter for the entailment category compared to the neutral category~\cite{gururangan2018annotation}. This suggests that crowd workers tend to remove words from the premise to create a hypothesis for the entailment category, leading to the overlap bias in the training data. Models trained on the skewed datasets will capture these artifacts and even amplify them during inference time.

\subsection{LLMs Models}\label{LLMs-models}
The robustness of NLU models is highly relevant to the pre-finetuned language models. In particular, there are two key factors: model sizes (measured by the number of parameters) and pre-training objectives.

First, models with the same kind of architectures and pre-training objective but with different sizes could have significantly different generalization ability. It is shown that increasing the size of the model could lead to an increase in the representation power and generalization ability. From the empirical perspective, comparisons have been made between LLMs of different sizes but with the same architecture, e.g., BERT-base with BERT-large, RoBERTa-base with RoBERTa-large~\cite{tu2020empirical,bhargava2021generalization}. The results show that the larger versions generalize consistently better than the base versions, with a relatively smaller accuracy gap between the OOD and IID test data. Smaller models have fewer parameters than larger models and have a smaller model capacity. Therefore, smaller models are more prone to capture spurious patterns and are more dependent on data artifacts for prediction. Another work~\cite{du2021compressed} studies the impact of model compression on the generalizability of LLMs and finds that compressed LLMs are significantly less robust compared to their uncompressed counterparts. Compressed models with knowledge distillation have also been shown to be more vulnerable to adversarial attacks. From a theoretical perspective, a recent analysis supports that there is a trade-off between the size of a model and its robustness, where large models tend to be more robust than smaller ones~\cite{bubeck2021universal}.

Second, LLMs with similar model sizes but with different pre-training objectives also differ in the generalization ability. Here, we consider three kinds of LLMs: BERT, ELECTRA, and RoBERTa. BERT is trained with masked language modeling and next-sentence prediction. RoBERTa removes the next-sentence prediction from BERT and uses dynamic masking. ELECTRA is trained to distinguish between real input tokens and fake input tokens generated by another network. Empirical analysis shows that these three models have significantly different levels of robustness~\cite{prasad2021extent}. For the Adversarial NLI (ANLI) dataset, it is shown that ELECTRA and RoBERTa have significantly better performance than BERT, for both the base and the large versions. Similarly, another study has shown that RoBERTa-base outperforms BERT-base around 20\% in terms of accuracy on the HANS test set~\cite{bhargava2021generalization}. 
Because different architectures have distinct object functions during the pre-training stage, different inductive biases may be encoded by the models. This could possibly explain their differences in generalizability.

\subsection{Model Fine-tuning Process}
The learning dynamics could reveal what knowledge has been learned during the course of model training. There are some observations. First, standard training procedures have a bias toward learning simple features, which we can refer to as the simplicity bias. The models are based mainly on the simplest features and remain invariant to complex predictive features. Moreover, it has been observed that the models give overconfident predictions for easy samples and low-confidence predictions for hard samples. Second, models tend to learn non-robust and easy-to-learn features at the early stage of training.
For example, reading comprehension models have learned the shortcut in the first few training iterations, which has influenced further exploration of the models for more robust features~\cite{lai2021machine}. Third, it has been experimentally validated that longer fine-tuning could lead to better generalization. Specifically, a larger number of training epochs will dramatically improve the generalizability of LLMs in NLU tasks~\cite{tu2020empirical}. 
The preference for non-robust features can be explained from the following perspective. The present LLM training methods can be considered as data-driven, corpus-based, statistical, and machine learning approaches. It is postulated that while this data-driven paradigm may prove effective in certain NLP tasks, it falls short in relevance to the challenging NLU tasks that necessitate a deeper understanding of natural language.

\section{Mitigation of Shortcut Learning}\label{sec:mitigation-approaches}
In this section, we introduce approaches that alleviate the problem of shortcut learning. The ultimate goal is to improve OOD generalization and adversarial robustness while still exhibiting good predictive performance in IID datasets. These methods are motivated mainly by the insights obtained in the last section. In particular, Section~\ref{Dataset-Refinement} introduces methods based on dataset refinement, and Section~\ref{Model-Centric} focuses on model-centric mitigation approaches.

\subsection{Data-Centric Mitigation Approaches}\label{Dataset-Refinement}

\vspace{3pt}
\noindent
\textbf{Dataset Refinement.}
Dataset refinement falls into the pre-processing mitigation family, with the aim of alleviating biases in the training datasets. First, when constructing new datasets, crowd workers will receive additional instructions to discourage the use of words that are highly indicative of annotation artifacts. Second, debiased datasets can also be developed by filtering out bias in existing data. For example, adversarial filtering is used to build a large-scale data set for the NLI task to reduce annotation artifacts that can be easily detected by a committee of strong baseline methods~\cite{zellers2018swag}. As a result, models trained on this dataset have to learn more generalizable features and rely on common sense reasoning to succeed. Third, we can also reorganize the train and test split, so that the bias distribution in the test set is different from that in the training set. Lastly, various types of data augmentation methods have been proposed. Representative examples include counterfactual data augmentation, mixup data augmentation, syntactically informative example augmentation by applying syntactic transformations to sentences, etc.

However, a drawback of this approach is that refining the dataset can only mitigate a limited number of recognized biases. The refined training set may not be completely free of biases and may still encompass statistical biases that are challenging for humans to identify. Thus, this could still negatively impact the model's performance.

\vspace{3pt}
\noindent
\textbf{Training Samples Reweighting.}
The main idea of reweighting is to place higher training weights on hard training samples, and vice versa~\cite{schuster2019towards,utama2020towards}. It is also called \emph{worst-group loss minimization} in some literature.
The underlying assumption is that improving the performance of the worst group (hard samples) is beneficial to the robustness of the model. It is typically achieved through two-stage training. In the first stage, the weight indexing model is trained; and in the second stage, the predictions of the indexing model are used as weights to adjust the importance of a training instance. Both soft weights~\cite{utama2020towards} and hard weights could be used in the second stage. Another representative example is focal loss, which is based on a regularizer to assign higher weights to hard samples that have less confident predictions.

\vspace{3pt}
\noindent
\textbf{Partitioning Data into Environments.}
This line of methods follows the principle of invariant risk minimization~\cite{arjovsky2019invariant}, which encourages models to learn invariants in multiple environments. For example, training data has been partitioned into several non-IID subsets (i.e., training environments), where spurious correlations vary across environments and reliable ones remain stable across environments~\cite{teney2020unshuffling}. The training scheme is designed to encourage the model to rely on stable correlations and suppress spurious correlations. Another work proposes an inter-environment matching objective by maximizing the inner product between gradients from different environments, with the goal of increasing model generalization~\cite{shi2021gradient}.

\subsection{Model-Centric Mitigation Methods}\label{Model-Centric}
In this section, we introduce model-centric mitigation methods, which can be named \textbf{robust learning} methods. These methods typically augment the traditional ERM-based training paradigm with different degrees of prior knowledge, explicitly or implicitly suppressing the model from capturing non-robust features. Some mitigation methods require that the shortcuts be known a priori, while others assume that the shortcuts are unknown.

\vspace{3pt}
\noindent
\textbf{Adversarial Training.}
This aims to learn better representations that do not contain information about artifacts or bias in the data. It is typically implemented in two ways in the NLP domain~\cite{stacey2020there,rashid2021mate}. First, the task classifier and adversarial classifier jointly share the same encoder~\cite{stacey2020there}. The goal of the adversarial classifier is to provide the correct predictions for the artifacts in the training data. Then the encoder and task classifier can be trained to optimize the task objective while reducing the performance of the adversarial classifier in predicting artifacts. Second, adversarial examples are generated to maximize a loss function, and the model is trained to minimize the loss function. For example, the generator based on the masked language model is used to perturb the text to generate adversarial samples~\cite{rashid2021mate}. Despite the difference, both methods leverage the MinMax formulation during the debiasing process.

\vspace{3pt}
\noindent
\textbf{Explanation Regularization.}
This category aims to regularize model training using prior knowledge established by humans~\cite{liu2019incorporating}.
Specifically, it is achieved by regularizing the feature attribution explanations with rationale annotations created by domain experts, to enforce the model to make the right predictions for the right reasons~\cite{liu2019incorporating}. These systems are trained to explicitly encourage the network to focus on features in the input that humans have annotated as important and suppress the models' attention to superficial patterns. For the NLI task, natural language explanations have been used to supervise the models, to encourage the model to pay more attention to the words present in the explanations~\cite{stacey2022supervising}. It has significantly improved the models' OOD generalization performance. Note that this type of method can only be used when prior knowledge is known in advance about shortcuts.

\vspace{3pt}
\noindent
\textbf{Product-of-Expert (PoE).}
The goal is to train a debiased model by training it as an ensemble with a bias-only model~\cite{clark2019don}.
This paradigm usually contains two stages. In the first stage, a bias-only model is explicitly trained to capture the bias of the data set, e.g. the hypothesis-only bias for the NLI task. During the second stage, the debiased model will be trained using cross-entropy loss, by combining its output with the output of the bias-only model:
$
\hat{p}_{i}=\operatorname{softmax}\left(\log \left(p_{i}\right)+\log \left(b_{i}\right)\right).
$
The parameters of the bias-only model are fixed during this stage, and only the debiased model parameters are updated by backpropagation. The goal is to encourage the debiased model to utilize orthogonal information with information from the bias-only model to make predictions.

\vspace{3pt}
\noindent
\textbf{Confidence Regularization.}
This mitigation scheme regularizes confidence in the model output, with the aim of encouraging the debiased model to give a higher uncertainty (lower confidence) for these biased samples. It is based on the observation that models tend to make overconfident predictions on biased examples. 
This relies on the training of a bias-only model to quantify the degree of bias of each training sample. The debiasing process is typically achieved through the knowledge distillation framework. In the first stage, the biased teacher model is trained using standard ERM loss, and the bias degree obtained from the bias-only model will be used to rescale the output distribution of the teacher model. In the second stage, the smoothed confidence values of the teacher model can be used to guide the training of the debiased model~\cite{du2021towards}.

\begin{figure*}
  \centering
  \includegraphics[width=0.93\linewidth]{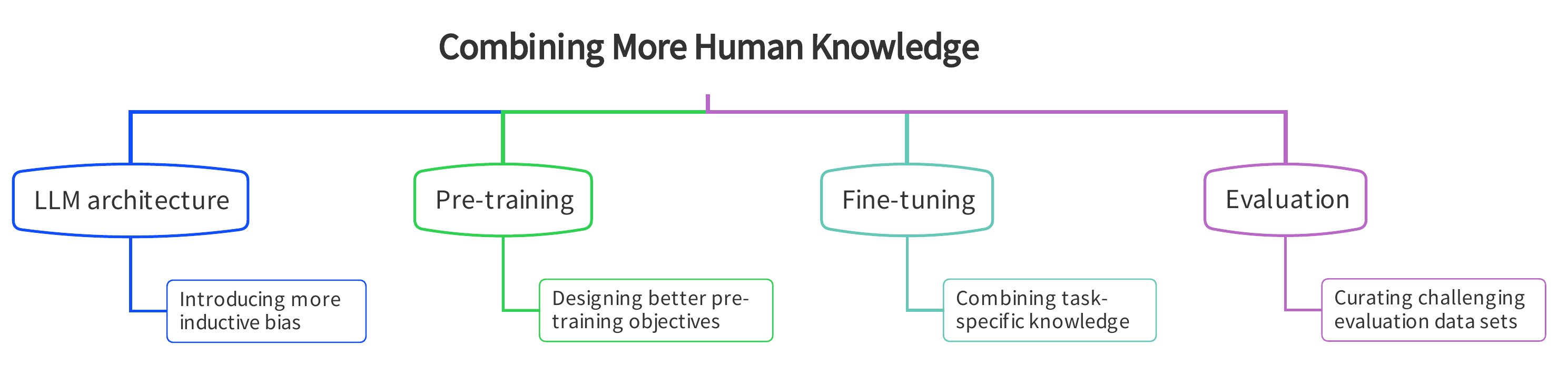}
  \caption{Combining more human knowledge to different stages of the pipeline. Specifically, knowledge can be combined to the architecture of model, model training process (both pre-training and fine-tuning), and model evaluation process.}
  \label{fig:combining-knowledge}
\end{figure*}

\vspace{3pt}
\noindent
\textbf{Contrastive Learning.}
Contrastive learning can be used to guide the training of representations. The goal is to construct the instance discrimination task to guide the model to capture the robust and predictive features, while suppressing the undesirable non-robust features. The instance discrimination task should be carefully designed; otherwise, it is possible to suppress robust predictive features~\cite{robinson2021can}.
A representative work presents a framework for mitigating spurious correlations using contrastive learning~\cite{choi2022c2l}. The method synthesizes a pair of factual and counterfactual inputs from the original text by masking identified causal and non-causal terms respectively. The model learns to associate the causal term with task labels by comparing the original text with its counterfactual counterpart, while learning to ignore non-causal features by contrasting with the factual pair. The framework leads to models that are less dependent on non-robust features and exhibit improved generalization performance.

\subsection{IID and Robustness Trade-off?}
Another open question is about the connection between IID performance and OOD robustness performance. To the best of our knowledge, there are no consistent observations. For example, there is a linear correlation between IID performance and OOD generalization for different types of models introduced in Section~\ref{LLMs-models}. On the contrary, most robust learning methods introduced in Section~\ref{Model-Centric} will sacrifice IID performance, although some of them could preserve IID performance. It deserves further research on the conditions under which the trade-off would occur. These insights could help the research community design robust learning frameworks that can simultaneously improve OOD and IID performance.

\section{Future Research Directions}
Despite the progress described in the previous sections, there are still numerous research challenges. In this section, we discuss potential research directions that could be pursued by the community.

\subsection{Introducing More Domain Knowledge}
Current standard of LLM training is data-driven. This is problematic because the resulting models essentially perform low-level pattern recognition. It may be useful for low-level NLP tasks like named-entity recognition (NER), but it is nearly impossible to tackle the more difficult natural language understanding tasks. As a result, it is preferable to combine the data-driven scheme with domain knowledge by incorporating knowledge at various stages of training. Furthermore, more knowledge should be applied to the design of the model architecture and the model evaluation (see Figure~\ref{fig:combining-knowledge}).

\vspace{3pt}
\noindent
\textbf{Inductive Bias to LLMs Models.}
It is suggested to introduce more inductive bias into the model architecture to improve robustness and generalization beyond IID benchmark datasets. Recently, some work has begun to induce certain kinds of linguistic structure in neural architectures. For example, TableFormer is proposed for robust table understanding~\cite{yang2022tableformer}. It proposes a structurally aware table-text encoding architecture, where tabular structural biases are incorporated through learnable attention biases. 
Although introducing linguistic-oriented biases to the model architectures might not result in the best performance for benchmark datasets, it is essential to improve generalization beyond IID benchmarks. 
Note that inductive biases are highly task-dependent and should be carefully designed for each specific task to accommodate its unique characteristic.

\vspace{3pt}
\noindent
\textbf{Better Pre-training Objectives.}
The pre-training objective also plays a crucial role in determining the OOD robustness of fine-tuned language models. As an example, recent studies have shown that pre-trained BERT embeddings suffer from strong anisotropy, meaning the average cosine similarity is significantly higher than zero and word vectors cluster in narrow cones in the vector space~\cite{liang2021learning,ethayarajh2019contextual}. This leads to word representations having a high similarity to unrelated words, impacting their expressive power and accuracy in downstream tasks.
It is desirable to invest more effort in designing better pre-training objectives to improve model robustness. 
Recent studies indicate that choosing a better pre-trained model could bring much better generalization performance than robust learning methods as introduced in Section~\ref{sec:mitigation-approaches}. For example, RoBERTa-base with a standard fine-tuning loss could even outperform the BERT-base with robust learning objectives in terms of generalization performance on the HANS test set~\cite{bhargava2021generalization}. This highlights the importance of pre-training in NLU model generalization performance and calls for increased community efforts to improve pre-trained language models.

\vspace{3pt}
\noindent
\textbf{Better Fine-tuning Approaches.} 
NLU tasks may contain various types of bias, which are not fully known even by domain experts. This is distinct from the literature that works with the toy task (e.g., Colored MNIST~\cite{arjovsky2019invariant}), which typically contains a single type of bias and the bias is fully known. As a result, the majority of existing mitigation methods for NLU tasks rely on human prior knowledge heuristics. Some examples include: i) weak models are more prone to capture biases, ii) non-robust models tend to give overconfident predictions for easy samples, etc. Unfortunately, this prior knowledge can only identify a limited number of biases in the data. Although it is possible to reduce the use of some identified shortcuts, models may still use other shortcuts for prediction. This could explain why existing mitigation methods only provide a limited improvement in generalization. As a result, it is suggested to incorporate more human-like common sense knowledge into the model training.

\vspace{3pt}
\noindent
\textbf{Curating Challenging Evaluation Datasets.}
It is encouraging to see that some benchmark datasets for adversarial and OOD robustness have emerged. For example, adversarial GLUE is proposed for adversarial robustness evaluation, which contains 14 adversarial attack methods~\cite{wang2021adversarial}. 
Despite these recent advances, it is necessary to continue curating difficult evaluation datasets that cover a wider range of NLU tasks, such as reading comprehension, and that cover a wider range of biases, such as those listed in Section~\ref{sec:bias-types}.

\subsection{Revisiting The Mitigation Approaches}
Existing mitigation methods have typically had limited mitigation performance. For example, for the MNLI task, the accuracy for mitigated models with BERT-base as the backbone is consistently lower than 70\% for the HANS test set~\cite{utama2020towards}. Note that HANS is a balanced binary test set, where 50\% is the accuracy of the random guess. The improvement in performance falls far short of our expectations. This brings up the following questions: 1) What have the mitigation algorithms accomplished, and 2) how can mitigation performance be improved further?

Debiased algorithms are thought to achieve better generalization because they can learn more robust features than biased models that rely primarily on non-robust features. However, this is not always the case with debiased algorithms. A recent work uses explainability as a debugging tool to analyze debiased models~\cite{mendelson2021debiasing}. The analysis indicates that the debiased models actually encode more biases in their inner representations. It is speculated that the improved performance on the OOD data comes from the refined classification head. More research is needed to investigate whether the debiased model has captured more robust features and what is the source of their improved generalization. 
This also suggests an interesting research direction by only updating the biased classification head, as updating the entire model is typically difficult and time consuming.

\subsection{In-depth Theoretical Understanding}
In addition to the current empirical research, there is also a growing trend of preliminary theoretical research aimed at uncovering the shortcut learning behavior of DNN models~\cite{shah2020pitfalls,vardi2022gradient,morwani2023simplicity}. For instance, using one-hidden-layer neural networks as the base model, one theoretical work uncovers that neural networks tend to  exclusively rely on simplest and non-robust features, while remain invariant to other useful but more complex features~\cite{shah2020pitfalls}. This type of \emph{simplicity bias} is one of the primary causes of low OOD generalization and adversarial vulnerability. Another theoretical study has investigated the reason behind superficial correlations from the optimization perspective~\cite{vardi2022gradient}. By using a depth-2 ReLU network as an example, the study proposed the \emph{Gradient Starvation} phenomenon, which states that the gradient descent optimization methods tend to learn non-robust networks while slowing down the learning of robust and task-relevant features. Although these existing works provide insights into the reason of shortcut learning of shallow neural networks, there is still a lack of a solid theoretical understanding of why LLMs learn shortcuts. In the future, further research is needed to fully explain this tendency in the context of LLMs.

\subsection{Taking Inspiration from Other Directions}
In addition, we can take inspiration from other relevant directions to address the shortcut learning issue of LLMs.

\vspace{3pt}
\noindent\textbf{Domain Adaptation \& Generalization.}
The robust learning approaches that we have discussed in Section~\ref{sec:mitigation-approaches} are closely relevant to domain adaptation and domain generalization. The three directions share the similarity that the training and test sets are not from the same distribution, i.e., there is a certain distribution shift. However, the objective of robust learning is distinct from domain adaptation, which aims to generalize to a specific target domain.
In contrast, robust learning is closer to domain generalization, where both areas have the goal of generalizing over a range of unknown conditions. The NLP community can leverage the findings from the domain generalization area to design more robust learning methods for LLMs.

\vspace{3pt}
\noindent\textbf{Long-Tailed Classification.}
Long-tailed classification addresses the issue of long-tailed distributed data, in which the head class has a large number of training samples while the tail class has few. Shortcut learning can be treated as a special case of long-tailed classification, where easy samples correspond to the head class and hard samples represent the tail class. Some of the robust learning solutions (e.g., reweighting) in Section~\ref{sec:mitigation-approaches} share a similar philosophy with approaches to the long-tailed classification problem. Leveraging ideas from approaches to long-tailed classification could improve the robustness of LLMs even further.

\vspace{3pt}
\noindent\textbf{Algorithmic Discrimination.}
Shortcut learning could also lead to discrimination and unfairness in deep learning models. In contrast to the general bias captured by the models, the spurious patterns here usually correspond to societal biases in terms of humans (e.g., racial bias and gender bias)~\cite{du2020fairness}. Here, the models have associated the fairness-sensitive attributes (e.g., ZIP code and surname) with main prediction task labels (e.g., mortgage loan rejection). At the inference time, the model would amplify the bias and show discrimination towards certain demographic groups, e.g., African Americans.

\subsection{Motivating Other Directions}
We can also take advantage of the insights discussed above to motivate the development of other directions.

\vspace{3pt}
\noindent\textbf{Backdoor Attack.}
The previous sections focus on discussing the setting in which LLMs have unintentionally captured undesirable shortcuts.
However, the adversary can intentionally insert shortcuts into LLMs, which could be a potential security threat to the deployed LLMs. This is termed the backdoor attack (or poisoning/Trojan attack)~\cite{tang2020embarrassingly}. Backdoor attackers insert human-crafted easy patterns that serve as shortcuts during the model training process, explicitly encouraging the model to learn shortcuts.
Representative examples include modifying the style of text, adding shortcut unigrams such as double quotation marks, etc.

\vspace{3pt}
\noindent\textbf{Watermarking.}
Unlike malicious use of shortcut learning as the backdoor attack, shortcut learning can also be used for benign purposes. In particular, trigger patterns can be inserted as watermarks by model owners during the training phase to protect the IP of companies. When LLMs are used by unauthorized users, shortcuts in the format of trigger patterns can be used by the stakeholders to claim ownership of the models.

\section{Prompt-based Paradigm} 
In previous sections, we have explored the characterization of the shortcut learning problem in the pre-training and fine-tuning training paradigm of medium-sized language models (typically with less than a billion parameters). With the recent emergence of huge-sized language models (with billions of parameters) such as GPT-3 and T5, the prompt-based paradigm has evolved into a new training paradigm with distinct formats from the standard fine-tuning paradigm. Consider the example of prompt for GPT-3. Using natural language instructions and/or demonstration of a few tasks, the LLM can generate the desired output without the need for gradient updates or fine-tuning. In this section, we examine the robustness of prompt-based methods and then compare them with the traditional standard fine-tuning approach.

\subsection{Robustness of Prompt-based Methods}
There are two types of prompt-based paradigms: 1) prompt-based fine-tuning and 2) prompting without fine-tuning. Prompt-based fine-tuning aims to enable medium-sized language models like BERT or RoBERTa to be few-shot learners, and this still requires optimizing the model's parameters. On the other hand, prompting without fine-tuning is meant for huge-sized language models like GPT-3, where the parameters are fixed and the model is applied to various tasks using different prompts, either discrete or soft. In the following discussion, we will discuss the shortcut learning issue in both types of prompt-based paradigms.

\vspace{3pt}
\noindent\textbf{Prompt-based Fine-tuning.} Preliminary research has been conducted to examine the shortcut learning challenge in the few-shot prompt-based fine-tuning paradigm~\cite{utama2021avoiding}. This preliminary study investigated the RoBERTa-large model, which comprises 355 million parameters. This work reveals the following insights: (i) zero-shot prompt-based models exhibit a higher level of robustness against the lexical overlap heuristic during inference, as evidenced by their strong performance on relevant challenge datasets. (ii) conversely, prompt-based finetuned models tend to adopt the spurious heuristic as they learn from larger amounts of labeled data, which is reflected by poor performance on OOD datasets. This indicates that prompt-based fine-tuning negatively impacts the robustness and generalizability of a model, just like the standard fine-tuning. The primary reason is that both training methods require adjusting the model's parameters using a biased NLI dataset, leading to a model that heavily relies on dataset biases as shortcuts for predictions.

\vspace{3pt}
\noindent\textbf{Prompting Without Fine-tuning.} Preliminary studies are emerging to examine the robustness of prompt-based methods for huge-size language models~\cite{webson2022prompt,zhao2021calibrate}. A study examines the few-shot learning performance of GPT-3 (2.7B, 13B, and 175B parameters) and GPT-2 (1.5B parameters) on text classification and information extraction tasks~\cite{zhao2021calibrate}. The results of the analysis reveal that the investigated LLMs are susceptible to majority label bias and position bias, where they tend to predict answers based on the frequency or position of the answers in the training data. Additionally, these LLMs also exhibit common token bias, where they favor answers that are prevalent in their pre-training corpus. Another study explores the impact of prompts on natural language inference tasks in zero-shot and few-shot settings using T0 (3B and 11B parameters) and GPT-3 (175B parameters)~\cite{webson2022prompt}. Experimental results suggest that models can learn just as quickly with many irrelevant or even misleading prompts as they can with effective and instructive prompts. This indicate that models’ improvement is not derived from models understanding task instructions in ways analogous to humans’ use of task instructions.

\subsection{Prompting versus Standard Fine-tuning}
\noindent
GPT-3's few-shot prompt performance is compared to that of BERT and RoBERTa through standard fine-tuning on two natural language inference tasks, i.e., MNLI and QQP~\cite{si2022prompting}. Additionally, these models are evaluated on the corresponding difficult OOD datasets: HANS and PAWS. The results show that GPT-3 performs slightly worse in generalization than BERT and RoBERTa on the in distribution MNLI and QQP datasets. On the other hand, GPT-3 achieves higher accuracy on the OOD tests for the majority of testing settings, indicating that GPT-3 has a lower generalization gap between the in distribution test set and the OOD test set, and thus a higher robustness. However, further analysis of the HANS dataset reveals that GPT-3 still exhibits substantial performance disparities between the bias-supporting and bias-countering subsets. This implies that there is room for enhancing the robustness of prompt-based techniques.

\vspace{3pt}
Note that the current research on prompt-based methods primarily aims at improving LLMs' performance on standard benchmarks. The robustness and generalization of this paradigm still require further investigation. A more thorough evaluation of prompt-based methods is needed and could be a future research topic. Additionally, techniques such as Chain-of-Thought~\cite{weichain} and Scratchpad~\cite{nyeshow} have been utilized to encourage models to perform intermediate calculations. These methods have proven to enhance the reasoning abilities of LLMs, thus having the potential to improve their robustness and generalization capabilities. Lastly, developing mitigation frameworks that can improve generalization performance on OOD test sets without sacrificing standard benchmark performance deserves more attention from the research community.

\section{Conclusions}
We present a thorough survey of the LLM's shortcut learning issue for NLU tasks in this article. Our findings suggest that shortcut learning is caused by a skewed dataset, model architecture, and model learning dynamics. We also summarize the mitigation solutions that can be used to reduce shortcut learning and improve the robustness of LLMs. 
Furthermore, we discuss directions that merit additional research effort from the research community, as well as the connections between shortcut learning and other relevant directions. The key takeaways from this survey's analysis are that the current pure data-driven training paradigm for LLMs is insufficient for high-level natural language understanding.  
In the future, the data-driven paradigm should be combined with domain knowledge at every stage of model design and evaluation to advance the field of LLMs.

\bibliographystyle{ACM-Reference-Format}
\bibliography{sample-base}


\begin{thebibliography}{55}


\ifx \showCODEN    \undefined \def \showCODEN     #1{\unskip}     \fi
\ifx \showDOI      \undefined \def \showDOI       #1{#1}\fi
\ifx \showISBNx    \undefined \def \showISBNx     #1{\unskip}     \fi
\ifx \showISBNxiii \undefined \def \showISBNxiii  #1{\unskip}     \fi
\ifx \showISSN     \undefined \def \showISSN      #1{\unskip}     \fi
\ifx \showLCCN     \undefined \def \showLCCN      #1{\unskip}     \fi
\ifx \shownote     \undefined \def \shownote      #1{#1}          \fi
\ifx \showarticletitle \undefined \def \showarticletitle #1{#1}   \fi
\ifx \showURL      \undefined \def \showURL       {\relax}        \fi
\providecommand\bibfield[2]{#2}
\providecommand\bibinfo[2]{#2}
\providecommand\natexlab[1]{#1}
\providecommand\showeprint[2][]{arXiv:#2}

\bibitem[Arjovsky et~al\mbox{.}(2019)]%
        {arjovsky2019invariant}
\bibfield{author}{\bibinfo{person}{Martin Arjovsky}, \bibinfo{person}{L{\'e}on
  Bottou}, \bibinfo{person}{Ishaan Gulrajani}, {and} \bibinfo{person}{David
  Lopez-Paz}.} \bibinfo{year}{2019}\natexlab{}.
\newblock \showarticletitle{Invariant risk minimization}.
\newblock \bibinfo{journal}{\emph{arXiv preprint arXiv:1907.02893}}
  (\bibinfo{year}{2019}).
\newblock


\bibitem[Bhargava et~al\mbox{.}(2021)]%
        {bhargava2021generalization}
\bibfield{author}{\bibinfo{person}{Prajjwal Bhargava},
  \bibinfo{person}{Aleksandr Drozd}, {and} \bibinfo{person}{Anna Rogers}.}
  \bibinfo{year}{2021}\natexlab{}.
\newblock \showarticletitle{Generalization in NLI: Ways (Not) To Go Beyond
  Simple Heuristics}. In \bibinfo{booktitle}{\emph{Proceedings of the Second
  Workshop on Insights from Negative Results in NLP}}.
\newblock


\bibitem[Branco et~al\mbox{.}(2021)]%
        {branco-etal-2021-shortcutted-commonsense}
\bibfield{author}{\bibinfo{person}{Ruben Branco}, \bibinfo{person}{Ant{\'o}nio
  Branco}, \bibinfo{person}{João Silva}, {and} \bibinfo{person}{João
  Rodrigues}.} \bibinfo{year}{2021}\natexlab{}.
\newblock \showarticletitle{Shortcutted Commonsense: Data Spuriousness in Deep
  Learning of Commonsense Reasoning}. In \bibinfo{booktitle}{\emph{Proceedings
  of the 2021 Conference on Empirical Methods in Natural Language Processing
  (EMNLP)}}.
\newblock


\bibitem[Brown et~al\mbox{.}(2020)]%
        {brown2020language}
\bibfield{author}{\bibinfo{person}{Tom~B Brown}, \bibinfo{person}{Benjamin
  Mann}, \bibinfo{person}{Nick Ryder}, \bibinfo{person}{Melanie Subbiah},
  \bibinfo{person}{Jared Kaplan}, \bibinfo{person}{Prafulla Dhariwal},
  \bibinfo{person}{Arvind Neelakantan}, \bibinfo{person}{Pranav Shyam},
  \bibinfo{person}{Girish Sastry}, \bibinfo{person}{Amanda Askell},
  {et~al\mbox{.}}} \bibinfo{year}{2020}\natexlab{}.
\newblock \showarticletitle{Language models are few-shot learners}.
\newblock \bibinfo{journal}{\emph{Advances in Neural Information Processing
  Systems (NeurIPS)}} (\bibinfo{year}{2020}).
\newblock


\bibitem[Bubeck and Sellke(2021)]%
        {bubeck2021universal}
\bibfield{author}{\bibinfo{person}{S{\'e}bastien Bubeck} {and}
  \bibinfo{person}{Mark Sellke}.} \bibinfo{year}{2021}\natexlab{}.
\newblock \showarticletitle{A universal law of robustness via isoperimetry}.
\newblock \bibinfo{journal}{\emph{Advances in Neural Information Processing
  Systems (NeurIPS)}} (\bibinfo{year}{2021}).
\newblock


\bibitem[Choi et~al\mbox{.}(2022)]%
        {choi2022c2l}
\bibfield{author}{\bibinfo{person}{Seungtaek Choi}, \bibinfo{person}{Myeongho
  Jeong}, \bibinfo{person}{Hojae Han}, {and} \bibinfo{person}{Seung-won
  Hwang}.} \bibinfo{year}{2022}\natexlab{}.
\newblock \showarticletitle{C2l: Causally contrastive learning for robust text
  classification}. In \bibinfo{booktitle}{\emph{Proceedings of the AAAI
  Conference on Artificial Intelligence}}, Vol.~\bibinfo{volume}{36}.
  \bibinfo{pages}{10526--10534}.
\newblock


\bibitem[Clark et~al\mbox{.}(2019)]%
        {clark2019don}
\bibfield{author}{\bibinfo{person}{Christopher Clark}, \bibinfo{person}{Mark
  Yatskar}, {and} \bibinfo{person}{Luke Zettlemoyer}.}
  \bibinfo{year}{2019}\natexlab{}.
\newblock \showarticletitle{Don't Take the Easy Way Out: Ensemble Based Methods
  for Avoiding Known Dataset Biases}.
\newblock \bibinfo{journal}{\emph{Empirical Methods in Natural Language
  Processing (EMNLP)}} (\bibinfo{year}{2019}).
\newblock


\bibitem[Devlin et~al\mbox{.}(2019)]%
        {devlin2018bert}
\bibfield{author}{\bibinfo{person}{Jacob Devlin}, \bibinfo{person}{Ming-Wei
  Chang}, \bibinfo{person}{Kenton Lee}, {and} \bibinfo{person}{Kristina
  Toutanova}.} \bibinfo{year}{2019}\natexlab{}.
\newblock \showarticletitle{Bert: Pre-training of deep bidirectional
  transformers for language understanding}.
\newblock \bibinfo{journal}{\emph{North American Chapter of the Association for
  Computational Linguistics (NAACL)}} (\bibinfo{year}{2019}).
\newblock


\bibitem[Du et~al\mbox{.}(2021)]%
        {du2021towards}
\bibfield{author}{\bibinfo{person}{Mengnan Du}, \bibinfo{person}{Varun
  Manjunatha}, \bibinfo{person}{Rajiv Jain}, \bibinfo{person}{Ruchi Deshpande},
  \bibinfo{person}{Franck Dernoncourt}, \bibinfo{person}{Jiuxiang Gu},
  \bibinfo{person}{Tong Sun}, {and} \bibinfo{person}{Xia Hu}.}
  \bibinfo{year}{2021}\natexlab{}.
\newblock \showarticletitle{Towards Interpreting and Mitigating Shortcut
  Learning Behavior of NLU Models}.
\newblock \bibinfo{journal}{\emph{North American Chapter of the Association for
  Computational Linguistics (NAACL)}} (\bibinfo{year}{2021}).
\newblock


\bibitem[Du et~al\mbox{.}(2023)]%
        {du2021compressed}
\bibfield{author}{\bibinfo{person}{Mengnan Du}, \bibinfo{person}{Subhabrata
  Mukherjee}, \bibinfo{person}{Yu Cheng}, \bibinfo{person}{Milad Shokouhi},
  \bibinfo{person}{Xia Hu}, {and} \bibinfo{person}{Ahmed~Hassan Awadallah}.}
  \bibinfo{year}{2023}\natexlab{}.
\newblock \showarticletitle{Robustness Challenges in Model Distillation and
  Pruning for Natural Language Understanding}.
\newblock \bibinfo{journal}{\emph{The 17th Annual Meeting of the European
  chapter of the Association for Computational Linguistics (EACL)}}
  (\bibinfo{year}{2023}).
\newblock


\bibitem[Du et~al\mbox{.}(2020)]%
        {du2020fairness}
\bibfield{author}{\bibinfo{person}{Mengnan Du}, \bibinfo{person}{Fan Yang},
  \bibinfo{person}{Na Zou}, {and} \bibinfo{person}{Xia Hu}.}
  \bibinfo{year}{2020}\natexlab{}.
\newblock \showarticletitle{Fairness in deep learning: A computational
  perspective}.
\newblock \bibinfo{journal}{\emph{IEEE Intelligent Systems}}
  (\bibinfo{year}{2020}).
\newblock


\bibitem[Ethayarajh(2019)]%
        {ethayarajh2019contextual}
\bibfield{author}{\bibinfo{person}{Kawin Ethayarajh}.}
  \bibinfo{year}{2019}\natexlab{}.
\newblock \showarticletitle{How Contextual are Contextualized Word
  Representations? Comparing the Geometry of BERT, ELMo, and GPT-2 Embeddings}.
  In \bibinfo{booktitle}{\emph{Proceedings of the 2019 Conference on Empirical
  Methods in Natural Language Processing and the 9th International Joint
  Conference on Natural Language Processing (EMNLP-IJCNLP)}}.
  \bibinfo{pages}{55--65}.
\newblock


\bibitem[Gururangan et~al\mbox{.}(2018)]%
        {gururangan2018annotation}
\bibfield{author}{\bibinfo{person}{Suchin Gururangan}, \bibinfo{person}{Swabha
  Swayamdipta}, \bibinfo{person}{Omer Levy}, \bibinfo{person}{Roy Schwartz},
  \bibinfo{person}{Samuel~R Bowman}, {and} \bibinfo{person}{Noah~A Smith}.}
  \bibinfo{year}{2018}\natexlab{}.
\newblock \showarticletitle{Annotation artifacts in natural language inference
  data}.
\newblock \bibinfo{journal}{\emph{North American Chapter of the Association for
  Computational Linguistics (NAACL)}} (\bibinfo{year}{2018}).
\newblock


\bibitem[Han et~al\mbox{.}(2020)]%
        {han2020explaining}
\bibfield{author}{\bibinfo{person}{Xiaochuang Han}, \bibinfo{person}{Byron~C
  Wallace}, {and} \bibinfo{person}{Yulia Tsvetkov}.}
  \bibinfo{year}{2020}\natexlab{}.
\newblock \showarticletitle{Explaining Black Box Predictions and Unveiling Data
  Artifacts through Influence Functions}. In
  \bibinfo{booktitle}{\emph{Proceedings of the 58th Annual Meeting of the
  Association for Computational Linguistics (ACL)}}.
\newblock


\bibitem[Jin et~al\mbox{.}(2020)]%
        {jin2020bert}
\bibfield{author}{\bibinfo{person}{Di Jin}, \bibinfo{person}{Zhijing Jin},
  \bibinfo{person}{Joey~Tianyi Zhou}, {and} \bibinfo{person}{Peter Szolovits}.}
  \bibinfo{year}{2020}\natexlab{}.
\newblock \showarticletitle{Is bert really robust? a strong baseline for
  natural language attack on text classification and entailment}. In
  \bibinfo{booktitle}{\emph{Proceedings of the AAAI conference on artificial
  intelligence (AAAI)}}.
\newblock


\bibitem[Ko et~al\mbox{.}(2020)]%
        {ko2020look}
\bibfield{author}{\bibinfo{person}{Miyoung Ko}, \bibinfo{person}{Jinhyuk Lee},
  \bibinfo{person}{Hyunjae Kim}, \bibinfo{person}{Gangwoo Kim}, {and}
  \bibinfo{person}{Jaewoo Kang}.} \bibinfo{year}{2020}\natexlab{}.
\newblock \showarticletitle{Look at the First Sentence: Position Bias in
  Question Answering}. In \bibinfo{booktitle}{\emph{Empirical Methods in
  Natural Language Processing (EMNLP)}}.
\newblock


\bibitem[Lai et~al\mbox{.}(2021)]%
        {lai2021machine}
\bibfield{author}{\bibinfo{person}{Yuxuan Lai}, \bibinfo{person}{Chen Zhang},
  \bibinfo{person}{Yansong Feng}, \bibinfo{person}{Quzhe Huang}, {and}
  \bibinfo{person}{Dongyan Zhao}.} \bibinfo{year}{2021}\natexlab{}.
\newblock \showarticletitle{Why Machine Reading Comprehension Models Learn
  Shortcuts?}
\newblock \bibinfo{journal}{\emph{ACL Findings}} (\bibinfo{year}{2021}).
\newblock


\bibitem[Liang et~al\mbox{.}(2021)]%
        {liang2021learning}
\bibfield{author}{\bibinfo{person}{Yuxin Liang}, \bibinfo{person}{Rui Cao},
  \bibinfo{person}{Jie Zheng}, \bibinfo{person}{Jie Ren}, {and}
  \bibinfo{person}{Ling Gao}.} \bibinfo{year}{2021}\natexlab{}.
\newblock \showarticletitle{Learning to remove: Towards isotropic pre-trained
  BERT embedding}.
\newblock \bibinfo{journal}{\emph{Artificial Neural Networks and Machine
  Learning--ICANN 2021: 30th International Conference on Artificial Neural
  Networks, Bratislava, Slovakia, September 14--17, 2021, Proceedings, Part V
  30}} (\bibinfo{year}{2021}), \bibinfo{pages}{448--459}.
\newblock


\bibitem[Liu and Avci(2019)]%
        {liu2019incorporating}
\bibfield{author}{\bibinfo{person}{Frederick Liu} {and} \bibinfo{person}{Besim
  Avci}.} \bibinfo{year}{2019}\natexlab{}.
\newblock \showarticletitle{Incorporating priors with feature attribution on
  text classification}.
\newblock \bibinfo{journal}{\emph{57th Annual Meeting of the Association for
  Computational Linguistics (ACL)}} (\bibinfo{year}{2019}).
\newblock


\bibitem[Liu et~al\mbox{.}(2019)]%
        {liu2019roberta}
\bibfield{author}{\bibinfo{person}{Yinhan Liu}, \bibinfo{person}{Myle Ott},
  \bibinfo{person}{Naman Goyal}, \bibinfo{person}{Jingfei Du},
  \bibinfo{person}{Mandar Joshi}, \bibinfo{person}{Danqi Chen},
  \bibinfo{person}{Omer Levy}, \bibinfo{person}{Mike Lewis},
  \bibinfo{person}{Luke Zettlemoyer}, {and} \bibinfo{person}{Veselin
  Stoyanov}.} \bibinfo{year}{2019}\natexlab{}.
\newblock \showarticletitle{Roberta: A robustly optimized bert pretraining
  approach}.
\newblock \bibinfo{journal}{\emph{arXiv preprint arXiv:1907.11692}}
  (\bibinfo{year}{2019}).
\newblock


\bibitem[McCoy et~al\mbox{.}(2019)]%
        {mccoy2019right}
\bibfield{author}{\bibinfo{person}{R~Thomas McCoy}, \bibinfo{person}{Ellie
  Pavlick}, {and} \bibinfo{person}{Tal Linzen}.}
  \bibinfo{year}{2019}\natexlab{}.
\newblock \showarticletitle{Right for the wrong reasons: Diagnosing syntactic
  heuristics in natural language inference}.
\newblock \bibinfo{journal}{\emph{57th Annual Meeting of the Association for
  Computational Linguistics (ACL)}} (\bibinfo{year}{2019}).
\newblock


\bibitem[Mendelson and Belinkov(2021)]%
        {mendelson2021debiasing}
\bibfield{author}{\bibinfo{person}{Michael Mendelson} {and}
  \bibinfo{person}{Yonatan Belinkov}.} \bibinfo{year}{2021}\natexlab{}.
\newblock \showarticletitle{Debiasing Methods in Natural Language Understanding
  Make Bias More Accessible}. In \bibinfo{booktitle}{\emph{Proceedings of the
  2021 Conference on Empirical Methods in Natural Language Processing
  (EMNLP)}}.
\newblock


\bibitem[Morwani et~al\mbox{.}(2023)]%
        {morwani2023simplicity}
\bibfield{author}{\bibinfo{person}{Depen Morwani}, \bibinfo{person}{Jatin
  Batra}, \bibinfo{person}{Prateek Jain}, {and} \bibinfo{person}{Praneeth
  Netrapalli}.} \bibinfo{year}{2023}\natexlab{}.
\newblock \showarticletitle{Simplicity Bias in 1-Hidden Layer Neural Networks}.
\newblock \bibinfo{journal}{\emph{arXiv preprint arXiv:2302.00457}}
  (\bibinfo{year}{2023}).
\newblock


\bibitem[Niven and Kao(2019)]%
        {niven2019probing}
\bibfield{author}{\bibinfo{person}{Timothy Niven} {and}
  \bibinfo{person}{Hung-Yu Kao}.} \bibinfo{year}{2019}\natexlab{}.
\newblock \showarticletitle{Probing neural network comprehension of natural
  language arguments}.
\newblock \bibinfo{journal}{\emph{57th Annual Meeting of the Association for
  Computational Linguistics (ACL)}} (\bibinfo{year}{2019}).
\newblock


\bibitem[Nye et~al\mbox{.}(2022)]%
        {nyeshow}
\bibfield{author}{\bibinfo{person}{Maxwell Nye}, \bibinfo{person}{Anders~Johan
  Andreassen}, \bibinfo{person}{Guy Gur-Ari}, \bibinfo{person}{Henryk
  Michalewski}, \bibinfo{person}{Jacob Austin}, \bibinfo{person}{David Bieber},
  \bibinfo{person}{David Dohan}, \bibinfo{person}{Aitor Lewkowycz},
  \bibinfo{person}{Maarten Bosma}, \bibinfo{person}{David Luan},
  {et~al\mbox{.}}} \bibinfo{year}{2022}\natexlab{}.
\newblock \showarticletitle{Show Your Work: Scratchpads for Intermediate
  Computation with Language Models}.
\newblock \bibinfo{journal}{\emph{Deep Learning for Code Workshop}}
  (\bibinfo{year}{2022}).
\newblock


\bibitem[Pezeshkpour et~al\mbox{.}(2021)]%
        {pezeshkpour2021combining}
\bibfield{author}{\bibinfo{person}{Pouya Pezeshkpour}, \bibinfo{person}{Sarthak
  Jain}, \bibinfo{person}{Sameer Singh}, {and} \bibinfo{person}{Byron~C
  Wallace}.} \bibinfo{year}{2021}\natexlab{}.
\newblock \showarticletitle{Combining feature and instance attribution to
  detect artifacts}.
\newblock \bibinfo{journal}{\emph{arXiv preprint arXiv:2107.00323}}
  (\bibinfo{year}{2021}).
\newblock


\bibitem[Pham et~al\mbox{.}(2020)]%
        {pham2020out}
\bibfield{author}{\bibinfo{person}{Thang~M Pham}, \bibinfo{person}{Trung Bui},
  \bibinfo{person}{Long Mai}, {and} \bibinfo{person}{Anh Nguyen}.}
  \bibinfo{year}{2020}\natexlab{}.
\newblock \showarticletitle{Out of Order: How important is the sequential order
  of words in a sentence in Natural Language Understanding tasks?}
\newblock \bibinfo{journal}{\emph{arXiv preprint arXiv:2012.15180}}
  (\bibinfo{year}{2020}).
\newblock


\bibitem[Prasad et~al\mbox{.}(2021)]%
        {prasad2021extent}
\bibfield{author}{\bibinfo{person}{Grusha Prasad}, \bibinfo{person}{Yixin Nie},
  \bibinfo{person}{Mohit Bansal}, \bibinfo{person}{Robin Jia},
  \bibinfo{person}{Douwe Kiela}, {and} \bibinfo{person}{Adina Williams}.}
  \bibinfo{year}{2021}\natexlab{}.
\newblock \showarticletitle{To what extent do human explanations of model
  behavior align with actual model behavior?}. In
  \bibinfo{booktitle}{\emph{Proceedings of the Fourth BlackboxNLP Workshop on
  Analyzing and Interpreting Neural Networks for NLP}}.
\newblock


\bibitem[Qi et~al\mbox{.}(2021)]%
        {qi-etal-2021-mind}
\bibfield{author}{\bibinfo{person}{Fanchao Qi}, \bibinfo{person}{Yangyi Chen},
  \bibinfo{person}{Xurui Zhang}, \bibinfo{person}{Mukai Li},
  \bibinfo{person}{Zhiyuan Liu}, {and} \bibinfo{person}{Maosong Sun}.}
  \bibinfo{year}{2021}\natexlab{}.
\newblock \showarticletitle{Mind the Style of Text! Adversarial and Backdoor
  Attacks Based on Text Style Transfer}. In
  \bibinfo{booktitle}{\emph{Proceedings of the 2021 Conference on Empirical
  Methods in Natural Language Processing (EMNLP)}}.
\newblock


\bibitem[Raffel et~al\mbox{.}(2020)]%
        {raffel2020exploring}
\bibfield{author}{\bibinfo{person}{Colin Raffel}, \bibinfo{person}{Noam
  Shazeer}, \bibinfo{person}{Adam Roberts}, \bibinfo{person}{Katherine Lee},
  \bibinfo{person}{Sharan Narang}, \bibinfo{person}{Michael Matena},
  \bibinfo{person}{Yanqi Zhou}, \bibinfo{person}{Wei Li}, {and}
  \bibinfo{person}{Peter~J Liu}.} \bibinfo{year}{2020}\natexlab{}.
\newblock \showarticletitle{Exploring the Limits of Transfer Learning with a
  Unified Text-to-Text Transformer}.
\newblock \bibinfo{journal}{\emph{Journal of Machine Learning Research (JMLR)}}
  (\bibinfo{year}{2020}).
\newblock


\bibitem[Rashid et~al\mbox{.}(2021)]%
        {rashid2021mate}
\bibfield{author}{\bibinfo{person}{Ahmad Rashid}, \bibinfo{person}{Vasileios
  Lioutas}, {and} \bibinfo{person}{Mehdi Rezagholizadeh}.}
  \bibinfo{year}{2021}\natexlab{}.
\newblock \showarticletitle{MATE-KD: Masked Adversarial TExt, a Companion to
  Knowledge Distillation}. In \bibinfo{booktitle}{\emph{Proceedings of the 59th
  Annual Meeting of the Association for Computational Linguistics and the 11th
  International Joint Conference on Natural Language Processing (ACL)}}.
\newblock


\bibitem[Robinson et~al\mbox{.}(2021)]%
        {robinson2021can}
\bibfield{author}{\bibinfo{person}{Joshua Robinson}, \bibinfo{person}{Li Sun},
  \bibinfo{person}{Ke Yu}, \bibinfo{person}{Kayhan Batmanghelich},
  \bibinfo{person}{Stefanie Jegelka}, {and} \bibinfo{person}{Suvrit Sra}.}
  \bibinfo{year}{2021}\natexlab{}.
\newblock \showarticletitle{Can contrastive learning avoid shortcut solutions?}
\newblock \bibinfo{journal}{\emph{Advances in Neural Information Processing
  Systems (NeurIPS)}} (\bibinfo{year}{2021}).
\newblock


\bibitem[Saxon et~al\mbox{.}(2023)]%
        {saxon2023peco}
\bibfield{author}{\bibinfo{person}{Michael Saxon}, \bibinfo{person}{Xinyi
  Wang}, \bibinfo{person}{Wenda Xu}, {and} \bibinfo{person}{William~Yang
  Wang}.} \bibinfo{year}{2023}\natexlab{}.
\newblock \showarticletitle{PECO: Examining Single Sentence Label Leakage in
  Natural Language Inference Datasets through Progressive Evaluation of Cluster
  Outliers}. In \bibinfo{booktitle}{\emph{Proceedings of the 17th Conference of
  the European Chapter of the Association for Computational Linguistics}}.
  \bibinfo{pages}{3053--3066}.
\newblock


\bibitem[Schuster et~al\mbox{.}(2019)]%
        {schuster2019towards}
\bibfield{author}{\bibinfo{person}{Tal Schuster}, \bibinfo{person}{Darsh~J
  Shah}, \bibinfo{person}{Yun Jie~Serene Yeo}, \bibinfo{person}{Daniel
  Filizzola}, \bibinfo{person}{Enrico Santus}, {and} \bibinfo{person}{Regina
  Barzilay}.} \bibinfo{year}{2019}\natexlab{}.
\newblock \showarticletitle{Towards debiasing fact verification models}.
\newblock \bibinfo{journal}{\emph{Empirical Methods in Natural Language
  Processing (EMNLP)}} (\bibinfo{year}{2019}).
\newblock


\bibitem[Sen and Saffari(2020)]%
        {sen2020models}
\bibfield{author}{\bibinfo{person}{Priyanka Sen} {and} \bibinfo{person}{Amir
  Saffari}.} \bibinfo{year}{2020}\natexlab{}.
\newblock \showarticletitle{What do Models Learn from Question Answering
  Datasets?}. In \bibinfo{booktitle}{\emph{Proceedings of the 2020 Conference
  on Empirical Methods in Natural Language Processing (EMNLP)}}.
\newblock


\bibitem[Shah et~al\mbox{.}(2020)]%
        {shah2020pitfalls}
\bibfield{author}{\bibinfo{person}{Harshay Shah}, \bibinfo{person}{Kaustav
  Tamuly}, \bibinfo{person}{Aditi Raghunathan}, \bibinfo{person}{Prateek Jain},
  {and} \bibinfo{person}{Praneeth Netrapalli}.}
  \bibinfo{year}{2020}\natexlab{}.
\newblock \showarticletitle{The pitfalls of simplicity bias in neural
  networks}.
\newblock \bibinfo{journal}{\emph{Advances in Neural Information Processing
  Systems (NeurIPS)}} (\bibinfo{year}{2020}).
\newblock


\bibitem[Shi et~al\mbox{.}(2022)]%
        {shi2021gradient}
\bibfield{author}{\bibinfo{person}{Yuge Shi}, \bibinfo{person}{Jeffrey Seely},
  \bibinfo{person}{Philip~HS Torr}, \bibinfo{person}{N Siddharth},
  \bibinfo{person}{Awni Hannun}, \bibinfo{person}{Nicolas Usunier}, {and}
  \bibinfo{person}{Gabriel Synnaeve}.} \bibinfo{year}{2022}\natexlab{}.
\newblock \showarticletitle{Gradient matching for domain generalization}.
\newblock \bibinfo{journal}{\emph{International Conference on Learning
  Representations (ICLR)}} (\bibinfo{year}{2022}).
\newblock


\bibitem[Si et~al\mbox{.}(2022)]%
        {si2022prompting}
\bibfield{author}{\bibinfo{person}{Chenglei Si}, \bibinfo{person}{Zhe Gan},
  \bibinfo{person}{Zhengyuan Yang}, \bibinfo{person}{Shuohang Wang},
  \bibinfo{person}{Jianfeng Wang}, \bibinfo{person}{Jordan Boyd-Graber}, {and}
  \bibinfo{person}{Lijuan Wang}.} \bibinfo{year}{2022}\natexlab{}.
\newblock \showarticletitle{Prompting gpt-3 to be reliable}.
\newblock \bibinfo{journal}{\emph{arXiv preprint arXiv:2210.09150}}
  (\bibinfo{year}{2022}).
\newblock


\bibitem[Si et~al\mbox{.}(2019)]%
        {si2019does}
\bibfield{author}{\bibinfo{person}{Chenglei Si}, \bibinfo{person}{Shuohang
  Wang}, \bibinfo{person}{Min-Yen Kan}, {and} \bibinfo{person}{Jing Jiang}.}
  \bibinfo{year}{2019}\natexlab{}.
\newblock \showarticletitle{What does BERT Learn from Multiple-Choice Reading
  Comprehension Datasets?}
\newblock \bibinfo{journal}{\emph{arXiv preprint arXiv:1910.12391}}
  (\bibinfo{year}{2019}).
\newblock


\bibitem[Sinha et~al\mbox{.}(2021)]%
        {sinha2021masked}
\bibfield{author}{\bibinfo{person}{Koustuv Sinha}, \bibinfo{person}{Robin Jia},
  \bibinfo{person}{Dieuwke Hupkes}, \bibinfo{person}{Joelle Pineau},
  \bibinfo{person}{Adina Williams}, {and} \bibinfo{person}{Douwe Kiela}.}
  \bibinfo{year}{2021}\natexlab{}.
\newblock \showarticletitle{Masked language modeling and the distributional
  hypothesis: Order word matters pre-training for little}.
\newblock \bibinfo{journal}{\emph{Empirical Methods in Natural Language
  Processing (EMNLP)}} (\bibinfo{year}{2021}).
\newblock


\bibitem[Stacey et~al\mbox{.}(2022)]%
        {stacey2022supervising}
\bibfield{author}{\bibinfo{person}{Joe Stacey}, \bibinfo{person}{Yonatan
  Belinkov}, {and} \bibinfo{person}{Marek Rei}.}
  \bibinfo{year}{2022}\natexlab{}.
\newblock \showarticletitle{Supervising Model Attention with Human Explanations
  for Robust Natural Language Inference}.
\newblock \bibinfo{journal}{\emph{AAAI Conference on Artificial Intelligence
  (AAAI)}} (\bibinfo{year}{2022}).
\newblock


\bibitem[Stacey et~al\mbox{.}(2020)]%
        {stacey2020there}
\bibfield{author}{\bibinfo{person}{Joe Stacey}, \bibinfo{person}{Pasquale
  Minervini}, \bibinfo{person}{Haim Dubossarsky}, \bibinfo{person}{Sebastian
  Riedel}, {and} \bibinfo{person}{Tim Rockt{\"a}schel}.}
  \bibinfo{year}{2020}\natexlab{}.
\newblock \showarticletitle{Avoiding the Hypothesis-Only Bias in Natural
  Language Inference via Ensemble Adversarial Training}.
\newblock \bibinfo{journal}{\emph{Empirical Methods in Natural Language
  Processing (EMNLP)}} (\bibinfo{year}{2020}).
\newblock


\bibitem[Sundararajan et~al\mbox{.}(2017)]%
        {sundararajan2017axiomatic}
\bibfield{author}{\bibinfo{person}{Mukund Sundararajan}, \bibinfo{person}{Ankur
  Taly}, {and} \bibinfo{person}{Qiqi Yan}.} \bibinfo{year}{2017}\natexlab{}.
\newblock \showarticletitle{Axiomatic Attribution for Deep Networks}.
\newblock \bibinfo{journal}{\emph{International Conference on Machine Learning
  (ICML)}} (\bibinfo{year}{2017}).
\newblock


\bibitem[Tang et~al\mbox{.}(2020)]%
        {tang2020embarrassingly}
\bibfield{author}{\bibinfo{person}{Ruixiang Tang}, \bibinfo{person}{Mengnan
  Du}, \bibinfo{person}{Ninghao Liu}, \bibinfo{person}{Fan Yang}, {and}
  \bibinfo{person}{Xia Hu}.} \bibinfo{year}{2020}\natexlab{}.
\newblock \showarticletitle{An embarrassingly simple approach for trojan attack
  in deep neural networks}. In \bibinfo{booktitle}{\emph{Proceedings of the
  26th ACM SIGKDD International Conference on Knowledge Discovery \& Data
  Mining (KDD)}}.
\newblock


\bibitem[Teney et~al\mbox{.}(2020)]%
        {teney2020unshuffling}
\bibfield{author}{\bibinfo{person}{Damien Teney}, \bibinfo{person}{Ehsan
  Abbasnejad}, {and} \bibinfo{person}{Anton van~den Hengel}.}
  \bibinfo{year}{2020}\natexlab{}.
\newblock \showarticletitle{Unshuffling data for improved generalization}.
\newblock \bibinfo{journal}{\emph{arXiv preprint arXiv:2002.11894}}
  (\bibinfo{year}{2020}).
\newblock


\bibitem[Tu et~al\mbox{.}(2020)]%
        {tu2020empirical}
\bibfield{author}{\bibinfo{person}{Lifu Tu}, \bibinfo{person}{Garima Lalwani},
  \bibinfo{person}{Spandana Gella}, {and} \bibinfo{person}{He He}.}
  \bibinfo{year}{2020}\natexlab{}.
\newblock \showarticletitle{An empirical study on robustness to spurious
  correlations using pre-trained language models}.
\newblock \bibinfo{journal}{\emph{Transactions of the Association for
  Computational Linguistics (TACL)}} (\bibinfo{year}{2020}).
\newblock


\bibitem[Utama et~al\mbox{.}(2020)]%
        {utama2020towards}
\bibfield{author}{\bibinfo{person}{Prasetya~Ajie Utama},
  \bibinfo{person}{Nafise~Sadat Moosavi}, {and} \bibinfo{person}{Iryna
  Gurevych}.} \bibinfo{year}{2020}\natexlab{}.
\newblock \showarticletitle{Towards debiasing NLU models from unknown biases}.
\newblock \bibinfo{journal}{\emph{Empirical Methods in Natural Language
  Processing (EMNLP)}} (\bibinfo{year}{2020}).
\newblock


\bibitem[Utama et~al\mbox{.}(2021)]%
        {utama2021avoiding}
\bibfield{author}{\bibinfo{person}{Prasetya~Ajie Utama},
  \bibinfo{person}{Nafise~Sadat Moosavi}, \bibinfo{person}{Victor Sanh}, {and}
  \bibinfo{person}{Iryna Gurevych}.} \bibinfo{year}{2021}\natexlab{}.
\newblock \showarticletitle{Avoiding Inference Heuristics in Few-shot
  Prompt-based Finetuning}.
\newblock \bibinfo{journal}{\emph{Empirical Methods in Natural Language
  Processing (EMNLP)}} (\bibinfo{year}{2021}).
\newblock


\bibitem[Vardi et~al\mbox{.}(2022)]%
        {vardi2022gradient}
\bibfield{author}{\bibinfo{person}{Gal Vardi}, \bibinfo{person}{Gilad Yehudai},
  {and} \bibinfo{person}{Ohad Shamir}.} \bibinfo{year}{2022}\natexlab{}.
\newblock \showarticletitle{Gradient Methods Provably Converge to Non-Robust
  Networks}.
\newblock \bibinfo{journal}{\emph{arXiv preprint arXiv:2202.04347}}
  (\bibinfo{year}{2022}).
\newblock


\bibitem[Wang et~al\mbox{.}(2021)]%
        {wang2021adversarial}
\bibfield{author}{\bibinfo{person}{Boxin Wang}, \bibinfo{person}{Chejian Xu},
  \bibinfo{person}{Shuohang Wang}, \bibinfo{person}{Zhe Gan},
  \bibinfo{person}{Yu Cheng}, \bibinfo{person}{Jianfeng Gao},
  \bibinfo{person}{Ahmed~Hassan Awadallah}, {and} \bibinfo{person}{Bo Li}.}
  \bibinfo{year}{2021}\natexlab{}.
\newblock \showarticletitle{Adversarial GLUE: A Multi-Task Benchmark for
  Robustness Evaluation of Language Models}.
\newblock \bibinfo{journal}{\emph{Thirty-fifth Conference on Neural Information
  Processing Systems Datasets and Benchmarks Track (Round 2)}}
  (\bibinfo{year}{2021}).
\newblock


\bibitem[Webson and Pavlick(2022)]%
        {webson2022prompt}
\bibfield{author}{\bibinfo{person}{Albert Webson} {and} \bibinfo{person}{Ellie
  Pavlick}.} \bibinfo{year}{2022}\natexlab{}.
\newblock \showarticletitle{Do Prompt-Based Models Really Understand the
  Meaning of Their Prompts?}. In \bibinfo{booktitle}{\emph{Proceedings of the
  2022 Conference of the North American Chapter of the Association for
  Computational Linguistics: Human Language Technologies}}.
\newblock


\bibitem[Wei et~al\mbox{.}(2022)]%
        {weichain}
\bibfield{author}{\bibinfo{person}{Jason Wei}, \bibinfo{person}{Xuezhi Wang},
  \bibinfo{person}{Dale Schuurmans}, \bibinfo{person}{Maarten Bosma},
  \bibinfo{person}{Fei Xia}, \bibinfo{person}{Ed~H Chi},
  \bibinfo{person}{Quoc~V Le}, \bibinfo{person}{Denny Zhou}, {et~al\mbox{.}}}
  \bibinfo{year}{2022}\natexlab{}.
\newblock \showarticletitle{Chain-of-Thought Prompting Elicits Reasoning in
  Large Language Models}. In \bibinfo{booktitle}{\emph{Advances in Neural
  Information Processing Systems}}.
\newblock


\bibitem[Yang et~al\mbox{.}(2022)]%
        {yang2022tableformer}
\bibfield{author}{\bibinfo{person}{Jingfeng Yang}, \bibinfo{person}{Aditya
  Gupta}, \bibinfo{person}{Shyam Upadhyay}, \bibinfo{person}{Luheng He},
  \bibinfo{person}{Rahul Goel}, {and} \bibinfo{person}{Shachi Paul}.}
  \bibinfo{year}{2022}\natexlab{}.
\newblock \showarticletitle{TableFormer: Robust Transformer Modeling for
  Table-Text Encoding}.
\newblock \bibinfo{journal}{\emph{60th Annual Meeting of the Association for
  Computational Linguistics (ACL)}} (\bibinfo{year}{2022}).
\newblock


\bibitem[Zellers et~al\mbox{.}(2018)]%
        {zellers2018swag}
\bibfield{author}{\bibinfo{person}{Rowan Zellers}, \bibinfo{person}{Yonatan
  Bisk}, \bibinfo{person}{Roy Schwartz}, {and} \bibinfo{person}{Yejin Choi}.}
  \bibinfo{year}{2018}\natexlab{}.
\newblock \showarticletitle{Swag: A large-scale adversarial dataset for
  grounded commonsense inference}.
\newblock \bibinfo{journal}{\emph{Proceedings of the 2018 Conference on
  Empirical Methods in Natural Language Processing (EMNLP)}}
  (\bibinfo{year}{2018}).
\newblock


\bibitem[Zhao et~al\mbox{.}(2021)]%
        {zhao2021calibrate}
\bibfield{author}{\bibinfo{person}{Zihao Zhao}, \bibinfo{person}{Eric Wallace},
  \bibinfo{person}{Shi Feng}, \bibinfo{person}{Dan Klein}, {and}
  \bibinfo{person}{Sameer Singh}.} \bibinfo{year}{2021}\natexlab{}.
\newblock \showarticletitle{Calibrate before use: Improving few-shot
  performance of language models}. In \bibinfo{booktitle}{\emph{International
  Conference on Machine Learning}}. PMLR, \bibinfo{pages}{12697--12706}.
\newblock


\end{thebibliography}

\end{document}